\documentclass{article}

\usepackage[preprint]{neurips_2026}

\usepackage[utf8]{inputenc} 
\usepackage[T1]{fontenc}    
\usepackage{hyperref}       
\usepackage{url}            
\usepackage{booktabs}       
\usepackage{amsfonts}       
\usepackage{nicefrac}       
\usepackage{microtype}      
\usepackage{xcolor}         
\usepackage{amsmath}         
\usepackage{graphicx}         
\usepackage{multirow}         
\usepackage{subcaption}       

\title{Learning Temporal Evidence Selection for Video Anomaly Understanding}

\author{%
Hangil Park\thanks{Equal contribution.}\quad
Seungheon Kim\footnotemark[1]\quad
Tae-Kyun Kim\thanks{Corresponding author.}\\[4pt]
Korea Advanced Institute of Science and Technology (KAIST)\\
\texttt{\{hangil.park, shkseiji5500, kimtaekyun\}@kaist.ac.kr}%
}

\begin{document}

\maketitle

\begin{abstract}
Video anomaly understanding (VAU) aims to generate language-level interpretations of anomalous events in long videos, requiring visual evidence that supports explanation beyond anomaly detection alone. Existing VAU pipelines commonly rely on anomaly-centric or predefined temporal sampling strategies, which efficiently emphasize suspicious moments but overlook contextual cues, object interactions, and temporally distributed evidence needed for reasoning. Motivated by recent progress in learnable temporal selection for long-video question answering, we revisit VAU temporal sampling as a downstream-aligned evidence selection problem. We introduce the semantic-conditioned Temporal Sampling Q-Former that learns to select temporal visual evidence through the VAU generation objective. Since generic VAU prompts provide weak video-specific guidance, we condition the sampler on CLIP-embedded object tags automatically extracted from the input video. These tags are used only for temporal evidence selection and are not appended to the LLM generation prompt, separating sampler conditioning from prompt augmentation. The proposed sampler operates in visual-token space, computes soft temporal selection weights from frame-level representations, and merges duplicated selections before response generation. Experiments on HIVAU-70K show that our model achieves the best performance among the compared fine-tuned models,  while ablations demonstrate that learnable temporal sampling improves over fixed anomaly-focused sampling and that CLIP-embedded object-tag conditioning further improves over generic-prompt conditioning. Our results suggest that temporal sampling in VAU should be aligned with explanation-oriented evidence utility rather than anomaly saliency alone.
\end{abstract}

\section{Introduction}
Video anomaly understanding (VAU) extends conventional video anomaly detection beyond deciding whether an anomaly occurs to explaining what happens and why it is anomalous. This shift changes what constitutes useful visual evidence. Whereas video anomaly detection~\citep{sultani2018real,tian2021rtfm,feng2021mist,li2022msl,chen2023mgfn} often rely on highly abnormal moments for scoring or localization, VAU frequently requires broader evidence—such as scene context~\citep{cuebench, cuva, pereira2026finevau}, object roles~\citep{pereira2026finevau, ma2025sherlock, zhou2026targetvau}, and temporally distributed observations~\citep{holmesvau, vader, vadtree}—that makes an event explainable in natural language. In other words, the evidence most useful for detecting abnormality is not necessarily the evidence most useful for explaining it. This distinction becomes especially important in long-video anomaly understanding, where only a small portion of frames contributes meaningfully to anomaly interpretation while the rest may be redundant or distracting for downstream reasoning~\citep{va-gpt}.

While recent VAU studies have improved anomaly understanding through multimodal reasoning, relation modeling, prompting, and other designs, we focus on temporal selection, which directly controls the visual evidence available to the language model. Holmes-VAU~\citep{holmesvau} introduces Anomaly Temporal Sampler (ATS) on top of the HIVAU-70K benchmark, VADER~\citep{vader} proposes Context-AwarE Sampling (CAES) together with causal relation modeling, and VA-GPT~\citep{va-gpt} further studies anomaly-oriented token selection for anomaly understanding. 
Although these methods differ in detail, their temporal selection criteria are often specified by predefined signals, such as anomaly scores, anomaly-aware heuristics, context priors, or externally derived temporal structures. Such proxy-guided selection is effective for emphasizing suspicious moments, but downstream anomaly reasoning often requires evidence beyond the most salient abnormal snippets.

\begin{figure*}[t]
    \centering
    \includegraphics[width=\textwidth]{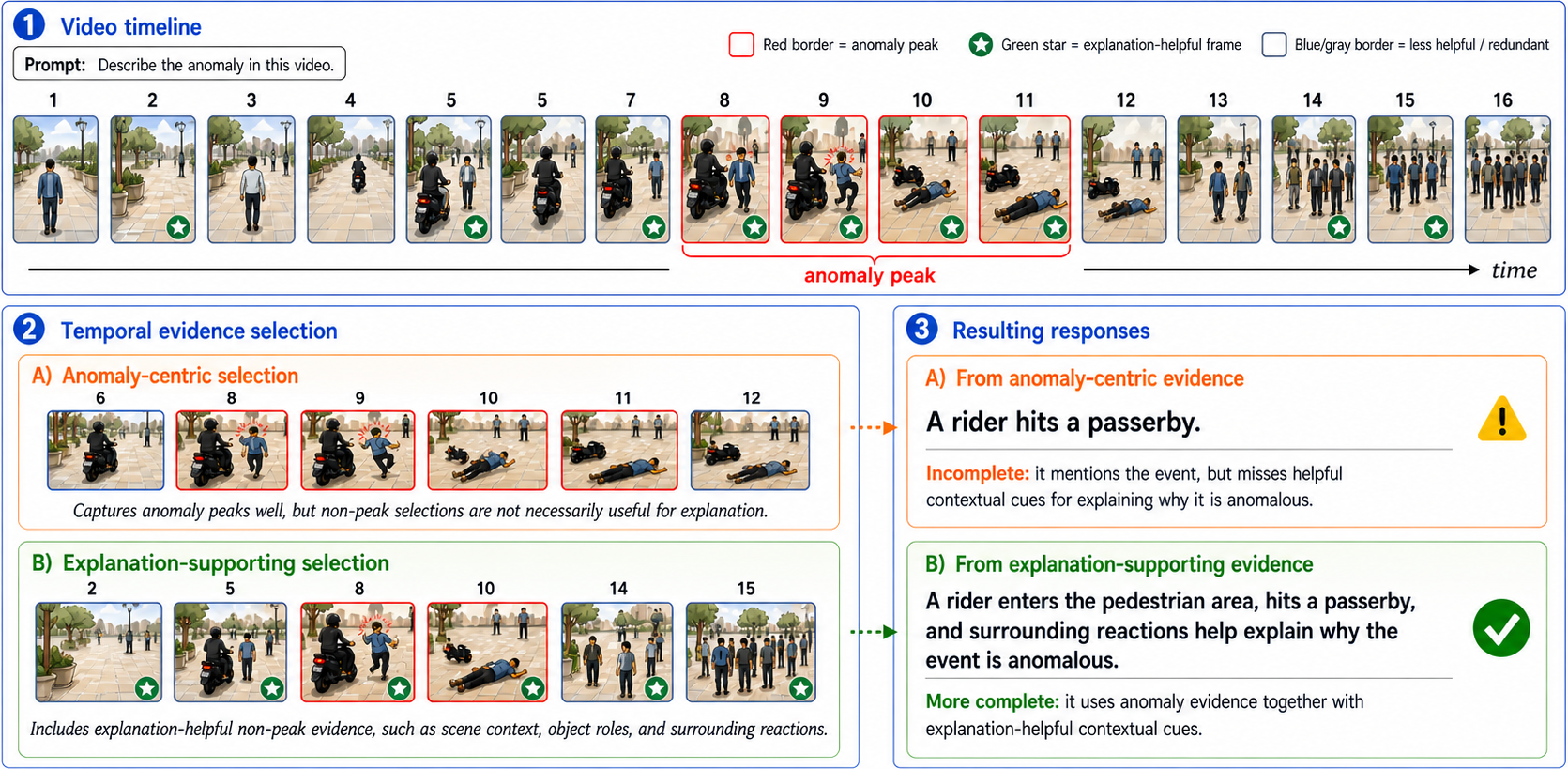}

  \caption{
\textbf{Conceptual illustration of the evidence mismatch in video anomaly understanding.} Anomaly-centric selection captures salient anomaly peaks, but its non-peak selections are not necessarily useful for explaining why the event is anomalous. Explanation-supporting selection retrieves anomaly evidence together with explanation-helpful non-peak cues, such as scene context, object roles, and surrounding reactions, leading to a more complete natural-language explanation.
}
\vspace{-5mm}
\label{fig:intro_motivation}
\end{figure*}

This mismatch makes temporal selection a central bottleneck in VAU. Rather than simply emphasizing the most anomalous moments, an effective sampler should preserve the temporally informative evidence needed for downstream reasoning across the video. As illustrated in Fig.~\ref{fig:intro_motivation}, anomaly-centered sampling  highlights suspicious regions, yet still fails to retain frames that are semantically important for explaining the event, such as contextual cues, object interactions, or temporally separated evidence. This observation suggests that temporal selection in VAU should be aligned with the downstream reasoning objective, rather than guided only by anomaly-side prioritization.

Video question answering (VQA) and video reasoning have recently studied temporal evidence selection as a core mechanism for scaling video-language models to long inputs. Since MLLMs can process only a limited number of visual tokens, these methods select or adapt frames according to downstream task relevance, user queries, or visual-token budgets, rather than relying only on uniform sampling~\citep{ffs,tang2025aks,yao2025generative,q-frame,bvllm}. For example, B-VLLM~\citep{bvllm} introduces text-conditioned adaptive frame selection and temporal merging to retain task-relevant spatio-temporal evidence under a limited token budget. This line of work suggests that temporal selection can be treated as a learnable or query-adaptive component, rather than a fixed pre-processing rule.

However, directly importing such VQA-oriented selection strategies into VAU is nontrivial. In general video QA, the question is often video-specific and provides a discriminative retrieval target: it specifies the relevant object, action, event, or temporal relation to look for. VAU, in contrast, typically relies on broad and nearly fixed prompts such as ``Describe the anomaly in this video''. Such a prompt indicates the task type, but it does not specify whether the relevant evidence lies in object interactions, scene context, temporal consequences, temporally distant contextual cues, or their combination. As a result, a sampler conditioned only on the generic anomaly-description prompt lacks sufficient video-specific guidance, even if the sampler itself is learnable or query-adaptive. Therefore, while long-video VQA methods provide a strong methodological starting point for temporal evidence selection, VAU requires a more informative sampler-conditioning signal that remains scalable without manual per-video supervision.

To address this challenge, we propose a semantic-conditioned Temporal Sampling Q-Former for VAU.
Instead of selecting frames according to anomaly-side scores, our sampler learns to select temporally informative evidence under the downstream VAU generation objective.
To provide stronger video-specific guidance than a generic anomaly-description prompt, we condition the sampler on CLIP-embedded object tags automatically extracted from the input video.
These tags are used only for temporal evidence selection and are not appended to the LLM generation prompt.
We instantiate the proposed sampler in an InternVL2-2B-based VAU pipeline and train it jointly with the semantic projection layer, and LLM LoRA adapters.
The proposed design shifts temporal selection from anomaly-side prioritization toward evidence selection for downstream anomaly reasoning.
The main contributions are as follows:
\begin{itemize}
    \item We revisit temporal evidence selection in VAU and identify a mismatch between anomaly-centric sampling and explanation-oriented evidence needs.
    \item We introduce a semantic-conditioned Temporal Sampling Q-Former that learns temporal evidence selection under the downstream VAU generation objective.
    \item We show on HIVAU-70K that learnable temporal sampling improves over fixed anomaly-focused sampling, and that CLIP-embedded object-tag conditioning improves over generic-prompt conditioning.
\end{itemize}

\section{Related Works}
\label{sec:related_works}

\subsection{Video Anomaly Detection and Anomaly Understanding}
\label{sec:video_anomaly_detection_and_anomaly_understanding}
Video anomaly detection (VAD) aims to localize events that deviate from normal patterns in videos. Recent weakly supervised VAD methods~\citep{sultani2018real,tian2021rtfm,feng2021mist,li2022msl,chen2023mgfn} mainly detect anomalous segments in long untrimmed videos from video-level labels, improving anomaly localization through MIL-based learning, robust snippet selection, temporal modeling, and discriminative feature learning. Although these methods have significantly advanced temporal localization, their outputs are primarily anomaly scores or event boundaries.

With the emergence of vision-language models and multimodal large language models, video anomaly analysis has increasingly moved toward video anomaly understanding (VAU), where models are expected to interpret anomalous events beyond localization. Recent works explore this transition through several complementary directions: textualizing video evidence and generating explanations with VLMs/LLMs~\citep{zanella2024lavad,lv2024vadllama}, inducing rules or learned guiding questions for abnormality reasoning~\citep{yang2024anomalyruler,vera}, and building agentic MLLM frameworks with scene-aware planning, retrieval, reflection, and memory~\citep{panda}. Other VAU studies focus on descriptive and interactive understanding~\citep{holmesvau, tang2024hawk}, context-aware and causal reasoning~\citep{cuebench, cuva, vader}, or structured event extraction and grounding~\citep{ma2025sherlock,vagu-gts}. These diverse directions indicate that VAU is not defined by a single methodological axis, but by a broader shift from localization-oriented detection toward semantic, contextual, causal, and grounded interpretation. Among these directions, we focus on temporal evidence selection, which directly determines which visual information is provided for downstream anomaly reasoning.

\subsection{Temporal Evidence Selection for Anomaly Analysis}
\label{sec:temporal_evidence_selection_for_anomaly_analysis}
Several recent anomaly analysis methods explicitly select frames, keyframes, or visual tokens as temporal evidence for anomaly reasoning. Holmes-VAU~\citep{holmesvau} introduces the Anomaly Temporal Sampler (ATS), and TargetVAU~\citep{zhou2026targetvau} employs an Anomaly-focused Temporal Sampler (AFTS); both use predicted anomaly scores with density-aware sampling to select anomaly-rich or behaviorally informative frames for anomaly understanding. VADER~\citep{vader} proposes Context-AwarE Sampling (CAES), which selects keyframes based on anomaly scores and temporal dynamics before relation-aware causal reasoning. VA-GPT~\citep{va-gpt} addresses evidence selection at the token level by aligning spatially and temporally effective visual tokens with LLMs for anomaly summarization and localization. Although these methods differ in granularity, they commonly determine the retained evidence using anomaly-side scores, anomaly-aware token alignment, or related proxy signals. As a result, the selection process is guided by cues designed to indicate abnormality, rather than by a learnable sampler explicitly optimized to select evidence for downstream VAU reasoning.

Other works structure temporal evidence without relying primarily on anomaly-score-guided frame selection. VADTree~\citep{vadtree} constructs a hierarchical granularity-aware tree from generic event-boundary cues, enabling coarse-to-fine temporal organization and redundancy reduction. A-Guardian~\citep{cuva} uses question-conditioned evidence retrieval and iterative prompt refinement to support causation understanding. These methods show that temporal evidence can be shaped not only by anomaly scores, but also by event boundaries, questions, or prompt-level guidance. However, their evidence construction still depends on external temporal priors, pretrained selectors, or indirect conditioning signals, rather than a temporal sampler learned to align evidence selection with downstream anomaly reasoning.

\subsection{Temporal Evidence Selection for General VQA}
\label{sec:temporal_evidence_selection_for_general_vqa}
Recent long-video understanding and video question answering methods have studied temporal evidence selection as a way to make video-language models reason over long inputs under limited visual-token budgets. One line of work performs query- or prompt-aware frame selection without fully training a task-specific sampler. AKS~\citep{tang2025aks} selects keyframes by balancing prompt-frame relevance and temporal coverage, and Q-Frame~\citep{q-frame} performs query-aware frame selection with multi-resolution adaptation in a training-free manner. GenS~\citep{yao2025generative} further formulates frame selection as a generative sampling problem, using a lightweight VideoLLM to identify question-relevant frames from long videos. These methods show that temporal selection can be guided by downstream textual conditions, such as prompts, user queries, or answer-relevant evidence.

Another line of work integrates frame selection more directly into the video-language reasoning pipeline. FFS~\citep{ffs} introduces a learnable frame-selection policy that selects informative frames for downstream multimodal reasoning, while B-VLLM~\citep{bvllm} uses a text-conditioned adaptive frame selection module followed by temporal token merging to retain task-relevant spatio-temporal cues under a fixed token budget. These methods suggest that temporal selection can be learned or adapted as part of the reasoning model, rather than treated only as an external preprocessing rule. These works demonstrate the value of learnable or query-adaptive temporal selection for video-language reasoning. Our work adapts this principle to VAU, where generic anomaly prompts provide limited video-specific guidance and temporal selection therefore benefits from additional semantic conditioning.
\section{Method}
\label{sec:method}

\subsection{Problem Formulation}
\label{sec:problem_formulation}

We address Video Anomaly Understanding (VAU) as a conditional video-to-text generation problem.
Given an input video $V$ and a task prompt or question $q$, the model generates a textual response $y=\{w_i\}_{i=1}^{N}$.
Depending on the prompt, the response may correspond to an anomaly judgement, a video description, or an explanation of what happens and why it is anomalous.

A central challenge in VAU is to select visual evidence that is useful for generating the target response.
Instead of relying on a fixed anomaly-centric sampling rule, we introduce a learnable temporal evidence selector optimized through the downstream VAU generation objective.
Formally, the model predicts the target response with the autoregressive likelihood
\vspace{-1.5mm}
\begin{equation}
    p_{\theta}(y \mid V, q)
    =
    \prod_{i=1}^{N}
    p_{\theta}(w_i \mid w_{<i}, V, q),
\end{equation}
and is trained with the standard cross-entropy loss
\vspace{-1mm}
\begin{equation}
    \mathcal{L}_{\mathrm{VAU}}
    =
    - \sum_{i=1}^{N}
    \log p_{\theta}(w_i \mid w_{<i}, V, q).
    \label{eq:vau_loss}
\end{equation}

The temporal selector receives no frame-level evidence supervision; it is trained only through the downstream generation loss.

\subsection{Overall Architecture}
\label{sec:method_overview}

\begin{figure*}[t]
    \centering
    \includegraphics[width=\textwidth]{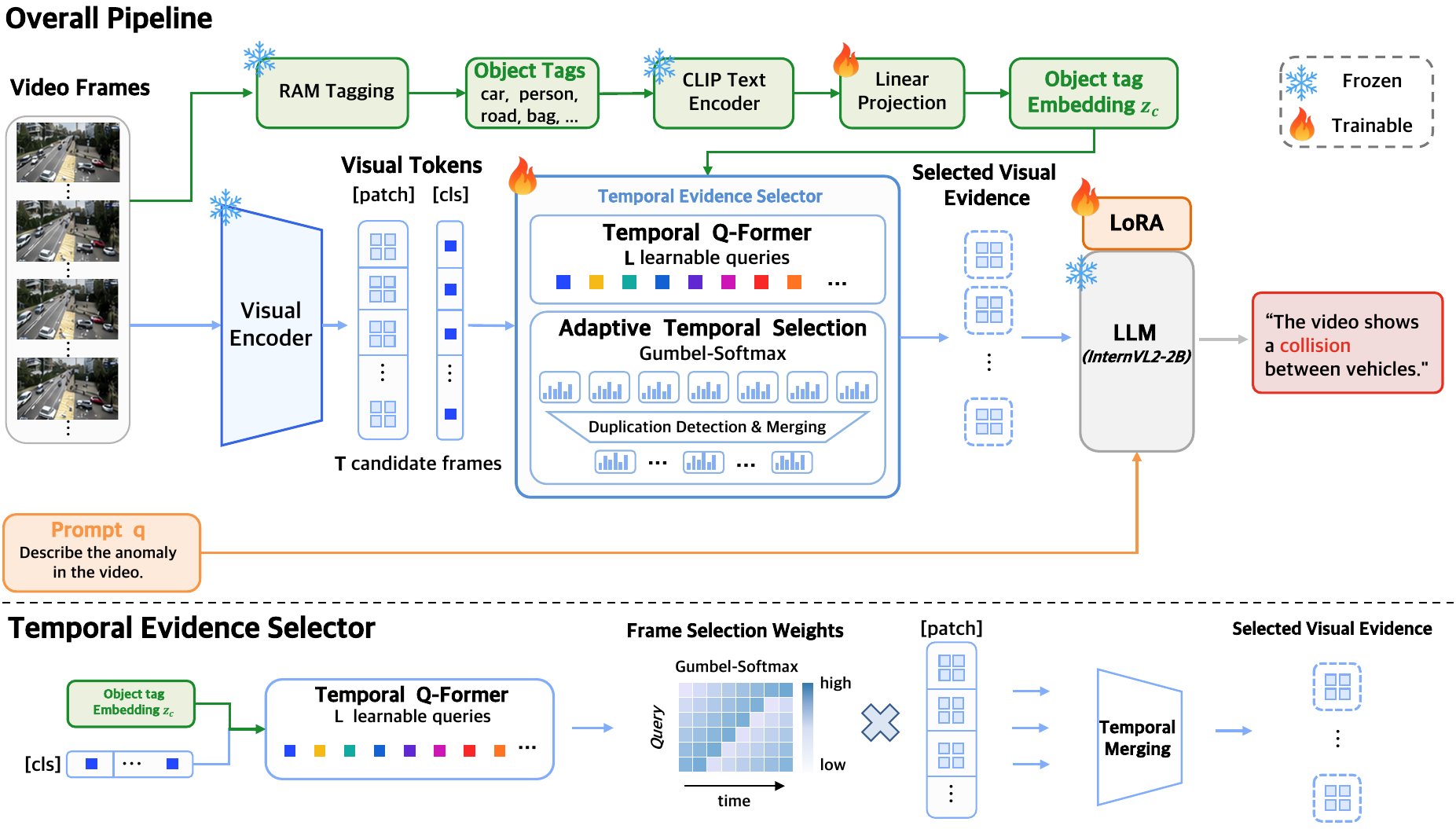}
    \caption{
    \textbf{Overview of the proposed temporal evidence selection framework.}
    The video is encoded by a frozen visual encoder, and the temporal Q-Former selects visual evidence from uniformly sampled candidate frames.
    Object-centric tags are encoded by a frozen CLIP text encoder and mapped to the temporal Q-Former hidden space by a trainable semantic projection layer.
    The resulting object-tag embedding sequence $\mathbf{Z}_c$ is used only to condition the temporal sampler.
    The selected evidence tokens are projected and then consumed by the LLM with the original prompt $q$.
    }
    \vspace{-3mm}
    \label{fig:method_overview}
\end{figure*}

Figure~\ref{fig:method_overview} illustrates the overall framework.
Our method builds on InternVL2-2B~\citep{internvl2} and replaces fixed anomaly-side temporal sampling with a learnable Temporal Q-Former trained through the downstream VAU generation objective.
The goal is to select temporal evidence that is useful for anomaly reasoning, rather than frames selected only by predefined anomaly-side scores or heuristics.

The key design choice is to condition the temporal evidence selector on video-specific object-centric semantic cues.
These cues are used only by the sampler and are not appended to the LLM generation prompt.


\subsection{Object-centric Semantic Conditioning}
\label{sec:semantic_conditioning}

Generic VAU prompts provide only weak video-specific guidance for temporal evidence selection.
To provide a more informative sampler condition, we extract object-centric tags from sparsely sampled frames using RAM~\citep{ram}.
Unlike prompt augmentation, these tags are used only to condition the temporal sampler and are not appended to the LLM generation prompt.

In our final model, each object tag is encoded by an external CLIP~\citep{clip} text encoder. 
Given a video-level tag set $\mathcal{T}=\{t_j\}_{j=1}^{M}$, we encode each tag independently:
\begin{equation}
    \mathbf{z}_j = \mathrm{CLIP}_{\mathrm{text}}(t_j),
    \qquad j=1,\ldots,M.
\end{equation}

Since the CLIP text embedding dimension differs from the temporal Q-Former hidden dimension, we apply a trainable semantic projection layer:
\begin{equation}
    \tilde{\mathbf{z}}_j = \mathbf{W}_{c}\mathbf{z}_j + \mathbf{b}_{c},
    \qquad j=1,\ldots,M.
\end{equation}
The projected tag embeddings are concatenated as a semantic conditioning sequence:
\begin{equation}
    \mathbf{Z}_c = [\tilde{\mathbf{z}}_1;\ldots;\tilde{\mathbf{z}}_M].
\end{equation}
The sequence $\mathbf{Z}_c$ is used as the semantic conditioning input to the temporal Q-Former, together with the frame-level CLS sequence used for temporal selection.
Implementation details of tag extraction and aggregation are provided in Appendix~\ref{app:semantic_hint_details}.

\subsection{Semantic-conditioned Temporal Sampling Q-Former}
\label{sec:temporal_selector}

Let $\{f_t\}_{t=1}^{T}$ denote uniformly sampled candidate frames. The frozen visual encoder produces frame-level visual tokens:
\begin{equation}
    \mathbf{X}_t =
    \{\mathbf{x}^{\mathrm{cls}}_t, \mathbf{x}_{t,1}, \ldots, \mathbf{x}_{t,P}\},
\end{equation}
where $\mathbf{x}^{\mathrm{cls}}_t$ is the frame-level CLS token and $\mathbf{x}_{t,p}$ denotes a patch-level token.
Following \citep{bvllm}, we use the frame-level CLS tokens as compact temporal representations for frame selection.
We denote the CLS sequence as
\vspace{-2mm}
\begin{equation}
    \mathbf{H}
    =
    [\mathbf{x}^{\mathrm{cls}}_1;\ldots;\mathbf{x}^{\mathrm{cls}}_T],
    \qquad
    \mathbf{H}\in\mathbb{R}^{T\times d}.
\end{equation}

The temporal Q-Former uses $L$ learnable temporal queries.
It jointly encodes the frame-level CLS sequence $\mathbf{H}$ and the semantic conditioning sequence $\mathbf{Z}_c$ to produce $L$ query representations:
\begin{equation}
    \mathbf{Q}
    =
    \mathrm{QFormer}_{\phi}
    (\mathbf{Q}_0;\mathbf{H},\mathbf{Z}_c),
    \qquad
    \mathbf{Q}\in\mathbb{R}^{L\times d},
\end{equation}
where $\mathbf{Q}_0\in\mathbb{R}^{L\times d}$ denotes the learnable temporal query tokens.
Each output query represents one temporal selection perspective.

We then compute temporal selection logits using scaled dot-product matching:
\begin{equation}
    \mathbf{A}
    =
    \frac{\tilde{\mathbf{Q}}\mathbf{H}^{\top}}{\sqrt{d}},
    \qquad
    \mathbf{A}\in\mathbb{R}^{L\times T},
\end{equation}
where \(\tilde{\mathbf{Q}}\) denotes the projected query representations.
We obtain soft temporal selection weights with Gumbel-Softmax:
\vspace{-1mm}
\begin{equation}
    \mathbf{S}_{\tau}
    =
    \mathrm{softmax}
    \left(
    \frac{\mathbf{A}+\mathbf{G}}{\tau}
    \right),
    \qquad
    \mathbf{G}\in\mathbb{R}^{L\times T}
    \text{ is i.i.d. Gumbel noise matrix}.
\end{equation}
Each row of $\mathbf{S}_{\tau}$ corresponds to one temporal query and defines a soft assignment over candidate frames.

The selection weights are computed from frame-level CLS tokens but are applied to the patch-level visual tokens that are forwarded to the LLM.
Let $\mathbf{x}_{t,p}$ denote the $p$-th retained visual token of frame $t$, where $p=1,\ldots,P$.
For the $\ell$-th temporal query, the selected evidence token is computed as
\vspace{-1mm}
\begin{equation}
    \mathbf{x}^{*}_{\ell,p}
    =
    \sum_{t=1}^{T}
    \mathbf{S}_{\tau}[\ell,t]\mathbf{x}_{t,p},
    \qquad
    \ell=1,\ldots,L,\; p=1,\ldots,P.
\end{equation}
This produces selected evidence tokens $\mathbf{X}^{*}\in\mathbb{R}^{L\times P\times d}$.
Thus, the sampler does not output raw frame indices for re-encoding; it directly produces selected temporal evidence in visual-token space.
We set $\tau=0.1$, which encourages sharp, nearly single-frame temporal assignments while retaining differentiable soft selection during training.

The temporal Q-Former produces $L$ temporal selection distributions
$\mathbf{S}_{\tau}\in\mathbb{R}^{L\times T}$ over the $T$ candidate frames.
We denote the $\ell$-th row by $\mathbf{s}_{\ell}=\mathbf{S}_{\tau}[\ell,:]$.
Since we use a low temperature $\tau=0.1$, each $\mathbf{s}_{\ell}$ is encouraged to be sharp and close to a single-frame assignment.
However, multiple temporal queries may still select the same or highly similar temporal evidence.
Following the temporal merging principle of~\citep{bvllm}, we detect duplicated selections using normalized row-wise cosine similarity between temporal selection distributions.
Specifically, two temporal queries $\ell$ and $k$ are treated as duplicated if
\begin{equation}
    \mathrm{sim}
    =
    \frac{1}{2}
    \left(
    1+
    \frac{s_l^\top s_k}{\|s_l\|\|s_k\|}
    \right)
    >
    \gamma,
\end{equation}
where $\gamma$ is the temporal merging threshold.
We apply this duplicate detection iteratively over the selected temporal queries and form groups of duplicated selections.
Let $\mathcal{G}_r$ denote the $r$-th group of duplicated temporal queries.

For each group $\mathcal{G}_r$, we merge the selected visual-token evidence by averaging the corresponding evidence tokens:
\vspace{-2mm}
\begin{equation}
    \bar{\mathbf{x}}_{r,p}
    =
    \frac{1}{|\mathcal{G}_r|}
    \sum_{\ell\in\mathcal{G}_r}
    \mathbf{x}^{*}_{\ell,p},
    \qquad
    p=1,\ldots,P,
\end{equation}
where $\mathbf{x}^{*}_{\ell,p}$ is the $p$-th selected visual token produced by the $\ell$-th temporal query.
Selections that are not assigned to any duplicate group are kept unchanged.
For each merged group, we average both the selected evidence tokens and their selection distributions. We define the representative temporal position as the argmax position of the merged selection distribution.
The retained evidence units are then ordered by their representative temporal positions and passed to the visual-token projector and LLM. This reduces redundant temporal evidence and produces at most $L$ retained evidence units.


\subsection{Training and Inference}
\label{sec:training_inference}

The selected and merged visual evidence tokens are passed through a visual-token projector and then consumed by the LLM together with the original prompt $q$.
The model generates the VAU response $y$ autoregressively.
We train the temporal Q-Former, the semantic projection layer, and LLM LoRA adapters using the VAU generation loss in Eq.~\eqref{eq:vau_loss}.
The visual encoder, RAM tagger, external CLIP text encoder, and base LLM parameters are frozen.

The temporal Q-Former is randomly initialized rather than initialized from a pretrained B-VLLM checkpoint.
Therefore, the selector learns to retrieve temporal evidence only through downstream VAU supervision.
Since no frame-level evidence labels are used, the method can be applied to VAU datasets that provide video-level prompts and textual responses but do not annotate which frames support the answer.
\section{Experiments}
\label{sec:experiments}

\subsection{Experimental Setup}
\label{sec:experimental_setup}

\paragraph{Dataset.}
We evaluate our method on HIVAU-70K, the hierarchical video anomaly understanding benchmark introduced by Holmes-VAU~\citep{holmesvau}.
We use the official train/test split and follow the clip-level, event-level, and video-level evaluation setting.

\paragraph{Evaluation metrics.}
We report BLEU~\citep{bleu}, CIDEr~\citep{cider}, METEOR~\citep{meteor}, and ROUGE~\citep{rouge} at three temporal granularities: clip-level (C), event-level (E), and video-level (V). Following prior VAU benchmarks~\citep{holmesvau, vader}, BLEU denotes the cumulative BLEU score, i.e., the sum of BLEU-1 to BLEU-4.

\paragraph{Compared methods.}
For the main benchmark, we compare reported HIVAU-70K results~\citep{holmesvau, vader} and our method using the same metrics.
We exclude VADER~\citep{vader} from the main comparison because it is not a sampler-only baseline: VADER combines CAES with relation-aware reasoning and external scene-graph/relation features for LLM training. Thus, its reported performance reflects additional relational feature extraction and generation modules beyond temporal evidence selection. To account for VADER's sampling component while keeping the comparison within our pipeline, we include a CAES sampler as a fixed inference-time baseline in our ablations. For the fixed-sampler baselines, we use a shared InternVL2-2B+LoRA checkpoint and change only the inference-time temporal sampling rule.
For the Temporal Q-Former variants, we train separate checkpoints end-to-end for each conditioning signal.
Across all ablation variants, we keep the backbone, LoRA configuration, training data and optimizer schedule.

\paragraph{Implementation details.}
For the learnable sampler, we uniformly sample at most $T = 64$ candidate frames from each video and use $L=12$ temporal queries. We use soft Gumbel-Softmax with $\tau=0.1$. We set temporal merging threshold $\gamma=0.9$ in all experiments. The temporal Q-Former, semantic projection layer, and LLM LoRA adapters are trainable, while the visual encoder, visual-token projector, RAM tagger, external CLIP text encoder, and base LLM parameters are frozen. Additional implementation details are provided in Appendix~\ref{app:implementation_details}.

\subsection{Main Results}
\label{sec:main_results}

Table~\ref{tab:main_results_hivau} compares our proposed method with generic zero-shot MLLMs and Holmes-VAU on HIVAU-70K. We group methods into zero-shot MLLMs and fine-tuned models, since the latter are fine-tuned with task-specific anomaly understanding data. 
Our method improves over Holmes-VAU across all reported metrics and temporal granularities. 
We use the ablations below to isolate the effect of learnable temporal sampling and sampler conditioning.

\begin{table*}[t]
\centering
\caption{
\textbf{Evaluation on the HIVAU-70K benchmark.} We report the performance at three temporal granularities: clip-level (C), event-level (E), and video-level (V). Ours denotes the Temporal Q-Former with Object Tags (CLIP Embedding), which is used as our final model. 
Rows marked with $^\dagger$ are evaluated or reproduced using our implementations; unmarked baseline results are from~\citep{holmesvau, vader}. 
The best results among the compared methods are shown in \textbf{bold}, and the second best results are \underline{underlined}.
}
\label{tab:main_results_hivau}
\resizebox{\textwidth}{!}{
\begin{tabular}{lc|ccc|ccc|ccc|ccc}
\toprule
\multirow{2}{*}{Method}
& \multirow{2}{*}{Params}
& \multicolumn{3}{c|}{BLEU~\citep{bleu}~($\uparrow$)}
& \multicolumn{3}{c|}{CIDEr~\citep{cider}~($\uparrow$)}
& \multicolumn{3}{c|}{METEOR~\citep{meteor}~($\uparrow$)}
& \multicolumn{3}{c}{ROUGE~\citep{rouge}~($\uparrow$)} \\
\cmidrule(lr){3-5}
\cmidrule(lr){6-8}
\cmidrule(lr){9-11}
\cmidrule(lr){12-14}
& 
& C & E & V
& C & E & V
& C & E & V
& C & E & V \\
\midrule
\multicolumn{14}{l}{\textit{Zero-shot MLLMs}} \\

Video-ChatGPT~\citep{video-chatgpt}
& 7B
& 0.152 & 0.068 & 0.066
& 0.033 & 0.011 & 0.013
& 0.102 & 0.069 & 0.044
& 0.153 & 0.048 & 0.079 \\

Video-LLAMA~\citep{video-llama}
& 7B
& 0.151 & 0.079 & 0.104
& 0.024 & 0.014 & 0.017
& 0.112 & 0.076 & 0.057
& 0.156 & 0.067 & 0.090 \\

Video-LLAVA~\citep{video-llava}
& 7B
& 0.164 & 0.046 & 0.055
& 0.032 & 0.009 & 0.013
& 0.097 & 0.022 & 0.014
& 0.132 & 0.023 & 0.045 \\

LLaVA-Next-Video~\citep{llava-video-next}
& 7B
& 0.435 & 0.091 & 0.120
& 0.102 & 0.015 & 0.031
& 0.117 & 0.085 & 0.096
& 0.198 & 0.080 & 0.106 \\

QwenVL2~\citep{qwen2}
& 7B
& 0.312 & 0.082 & 0.155
& 0.044 & 0.020 & 0.044
& 0.133 & 0.092 & 0.112
& 0.163 & 0.081 & 0.137 \\

InternVL2~\citep{internvl2}
& 8B
& 0.331 & 0.101 & 0.145
& 0.052 & 0.022 & 0.035
& 0.141 & 0.095 & 0.101
& 0.182 & 0.102 & 0.122 \\

NVILA~\citep{nvila}$^\dagger$
& 2B
& 0.528 & 0.176 & 0.210
& 0.090 & 0.044 & 0.057
& 0.131 & 0.112 & 0.103
& 0.199 & 0.152 & 0.150 \\

\midrule
\multicolumn{14}{l}{\textit{Fine-tuned models}} \\

Holmes-VAU~\citep{holmesvau} (InternVL2)
& 2B
& \underline{0.913} & \underline{0.804} & \underline{0.566}
& \underline{0.467} & \underline{1.519} & \underline{1.437}
& \underline{0.190} & \underline{0.165} & \underline{0.121}
& \underline{0.329} & \underline{0.370} & \underline{0.355} \\

Ours (InternVL2)$^\dagger$
& 2B

& \textbf{1.205} & \textbf{0.956} & \textbf{0.782}
& \textbf{0.566} & \textbf{1.575} & \textbf{1.525}
& \textbf{0.208} & \textbf{0.214} & \textbf{0.160}
& \textbf{0.360} & \textbf{0.420} & \textbf{0.399} \\


                                           
\bottomrule
\end{tabular}
}
\end{table*}
\subsection{Ablation Studies}
\label{sec:ablations}
Table~\ref{tab:pooled_ablations} reports ablation results computed directly on the full HIVAU-70K evaluation set, including clip-, event-, and video-level samples. The end-to-end trained Temporal Q-Former variants substantially improve BLEU and CIDEr over the fixed-sampler baselines. Within the Temporal Q-Former block, object-tag conditioning improves over generic-prompt conditioning, and the CLIP embedding variant achieves the best point estimate across all four metrics. We therefore use Object Tags (CLIP Embedding) as our final conditioning strategy. Since full evaluation set scores can hide granularity- and task-specific behavior, we provide C/E/V breakdowns and task-wise analyses in Appendix~\ref{app:additional_sampling}. To assess the stability of the pairwise differences, we provide paired bootstrap confidence intervals for all metrics in Appendix~\ref{app:paired_bootstrap}.

\begin{table*}[t]
\centering
\caption{
\textbf{Ablation study on temporal evidence selection and sampler conditioning.}
Scores are computed directly on the full HIVAU-70K evaluation set, including clip-, event-, and video-level samples.
The upper block compares fixed temporal sampling strategies, while the lower block compares Temporal Q-Former variants with different conditioning signals.
For the fixed-sampler baselines, we use a shared InternVL2-2B+LoRA checkpoint and change only the inference-time sampling rule.
For the Temporal Q-Former variants, we train separate checkpoints for each conditioning signal while keeping the backbone and LoRA training configuration.
}
\label{tab:pooled_ablations}
\resizebox{\textwidth}{!}{
\begin{tabular}{llcccc}
\toprule
Evidence Selector
& Conditioning Signal
& BLEU~($\uparrow$)
& CIDEr~($\uparrow$)
& METEOR~($\uparrow$)
& ROUGE~($\uparrow$) \\
\midrule
Uniform Sampling
& --
& 0.9457 & 0.8396 & 0.2027 & 0.3744 \\

ATS~\citep{holmesvau}
& --
& 0.9144 & 0.8537 & 0.2008 & 0.3738 \\

CAES~\citep{vader}
& --
& 0.9851 & 0.8582 & 0.1989 & 0.3690 \\

\midrule
Temporal Q-Former
& Generic Prompt
& 1.1227 & 0.9015 & 0.1996 & 0.3729 \\

Temporal Q-Former
& Object Tags (Text)
& 1.1274 & 0.9188 & 0.1996 & 0.3759 \\

Temporal Q-Former
& Object Tags (CLIP Embedding)
& \textbf{1.1482}
& \textbf{0.9334}
& \textbf{0.2028}
& \textbf{0.3776} \\


\bottomrule
\end{tabular}
}
\vspace{-5mm}
\end{table*}

\subsection{Qualitative Analysis}
\label{sec:qualitative_analysis}

\begin{figure*}[t]
    \centering
    \includegraphics[width=\textwidth]{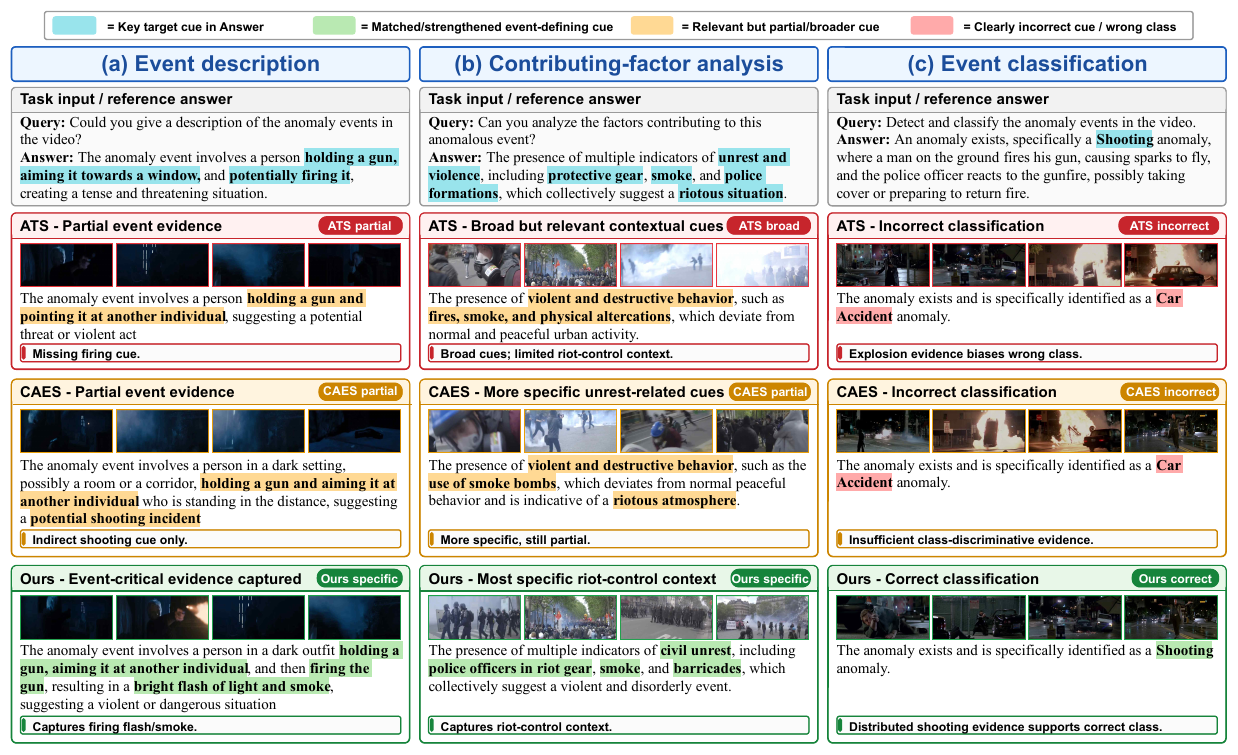}
\vspace{-5mm}
\caption{
\textbf{Qualitative comparison of temporal evidence selection on HIVAU-70K.}
We compare ATS, CAES, and the object-tag-conditioned Temporal Q-Former.
The proposed sampler retrieves explanation-supporting evidence missed by fixed anomaly-focused samplers, including brief causal moments and temporally distributed contextual cues needed for anomaly reasoning.
}
\vspace{-5mm}
\label{fig:qualitative_selection}
\end{figure*}

Figure~\ref{fig:qualitative_selection} provides a qualitative comparison of temporal evidence selection.
The figure compares the frames selected by ATS~\citep{holmesvau}, CAES~\citep{vader}, and the Temporal Q-Former with object-tag conditioning, together with the generated responses.
This analysis is intended to show whether the proposed sampler retrieves temporally distributed and semantically relevant evidence that supports the VAU response.

In the first example, ATS and CAES generate responses that are largely consistent with conservative ground-truth annotation, which describes the person as holding, aiming, or potentially firing a gun. However, this agreement is achieved without retaining the actual firing moment: their sampled frames cover gun aiming and smoke, but miss the brief gunfire frame that causally links the action to the observed smoke. As a result, the response remains at the level of ground-truth ambiguity, describing the event as ``potentially firing'' rather than explicitly identifying the gunshot. In contrast, our sampler captures the key firing frame, enabling the model to ground the description in the actual event and explain the smoke as an after-effect of the gunfire.
In the second example, ATS mainly samples visually intense moments such as fire, smoke, and physical altercations and therefore produces generic factors of violence and destruction. CAES captures a more relevant riot-related cue by mentioning smoke bombs and the riotous atmosphere, but its evidence remains incomplete and does not fully cover the broader scene context. In contrast, our sampler selects not only highly salient violent moments but also semantically important frames with lower anomaly saliency, including police officers in riot gear, formations, and barricades. These frames provide the model with a police-response and crowd-control context, enabling a more event-specific analysis of civil unrest rather than a generic description of violent behavior. 
In the third example, ATS and CAES mainly sample the local segment where a vehicle explodes after the shooting, because this visually dramatic moment receives the highest anomaly scores. Although these frames are indeed anomalous, they provide incomplete evidence about the broader event and bias the model toward the wrong class, leading it to predict a car accident. In contrast, our sampler selects temporally distributed frames that capture the gunfight occurring throughout the video, rather than focusing only on the explosion aftermath. By preserving evidence of armed confrontation before and around the explosion, our method enables the model to recover the correct anomaly class as shooting.

These examples show that anomaly-focused sampling can capture visually suspicious moments, but downstream-aligned temporal selection better preserves event-defining and contextually informative evidence for anomaly reasoning. For visualization clarity, we show only four frames out of the twelve sampled frames for each method; the full sampled-frame sequences are provided in Appendix~\ref{app:additional_qualitative}.
\vspace{-2mm}
\section{Conclusion}
\label{sec:conclusion}
\vspace{-2mm}
We studied temporal evidence selection for video anomaly understanding from the perspective of downstream explanation utility. Motivated by the mismatch between anomaly-centric sampling and explanation-oriented evidence needs, we introduced a semantic-conditioned Temporal Sampling Q-Former that learns to select temporal visual evidence through the VAU generation objective. The sampler is conditioned on CLIP-embedded object tags extracted from the input video, while the LLM receives only the original VAU prompt. Experiments on HIVAU-70K show that learnable temporal sampling improves over anomaly-focused sampling, and that object-tag conditioning further improves over generic-prompt conditioning on most metrics and temporal granularities. These findings suggest that VAU sampling should move beyond anomaly saliency alone and be designed around evidence useful for natural-language anomaly reasoning.

\paragraph{Limitations and broader impacts.}
We discuss limitations, future work, and broader impacts in Appendix~\ref{app:limitations_impacts}.

\newpage
{
\small
\bibliographystyle{unsrt}
\bibliography{reference}
}

\newpage
\appendix
\section{Limitations, Future Work, and Broader Impacts}
\label{app:limitations_impacts}

\paragraph{Limitations and future work.}
Our experiments are mainly conducted on HIVAU-70K, and further validation on additional VAU benchmarks would strengthen the generality of the findings. The proposed conditioning depends on off-the-shelf RAM and CLIP text encoders, which may miss fine-grained relations, fail to capture rare objects, or introduce recognition bias into the sampler. We also do not explicitly evaluate robustness to noisy or missing object tags, nor do we provide a systematic latency or FLOPs analysis of the additional RAM extraction, CLIP encoding, and temporal Q-Former components. Future work should explore end-to-end learnable semantic conditioning and more efficient evidence-selection pipelines that better align tag extraction, temporal sampling, and VAU generation. In addition, our task-aware extension experiment in Appendix~\ref{app:task_aware_extension} suggests that a simple global prompt-type tag is not sufficient to uniformly improve performance across clip-, event-, and video-level samples; developing granularity-aware or question-aware semantic conditioning remains future work.

\paragraph{Broader impacts.}
Video anomaly understanding can support more interpretable video analysis by providing language-level explanations of anomalous events rather than only anomaly scores. However, such systems may also be misused in surveillance or monitoring contexts, where incorrect explanations or biased visual evidence selection could affect human judgment. Our work is evaluated only on benchmark data and is not a deployed decision-making system. Practical deployment would require careful human oversight, privacy protection, bias evaluation, and domain-specific validation.

\section{Additional Sampling Analysis}
\label{app:additional_sampling}

\subsection{Granularity-wise Ablation Results}
\label{app:granularity_ablation}

Table~\ref{tab:appendix_sampling_ablation} and Table~\ref{tab:conditioning_ablation_full} provide C/E/V breakdowns corresponding to the full-set ablation results in Table~\ref{tab:pooled_ablations}.
These breakdowns show that the effect of sampling varies across temporal granularity and metric.

\begin{table*}[h]
\centering
\caption{
\textbf{C/E/V breakdown of sampling strategy ablations.} This table reports clip-, event-, and video-level results for the different sampling strategies.
}
\label{tab:appendix_sampling_ablation}
\resizebox{\textwidth}{!}{
\begin{tabular}{lcccccccccccc}
\toprule
\multirow{2}{*}{Sampling Strategy}
& \multicolumn{3}{c}{BLEU~($\uparrow$)}
& \multicolumn{3}{c}{CIDEr~($\uparrow$)}
& \multicolumn{3}{c}{METEOR~($\uparrow$)}
& \multicolumn{3}{c}{ROUGE~($\uparrow$)} \\
\cmidrule(lr){2-4}
\cmidrule(lr){5-7}
\cmidrule(lr){8-10}
\cmidrule(lr){11-13}
& C & E & V
& C & E & V
& C & E & V
& C & E & V \\
\midrule
Uniform Sampling
& 1.1297 & 0.6712 & 0.4999
& 0.5061 & 1.4220 & 1.3516
& \textbf{0.2151} & 0.1996 & 0.1420
& \textbf{0.3618} & 0.4059 & 0.3858 \\

ATS~\citep{holmesvau}
& 1.0937 & 0.6028 & 0.5348
& 0.5037 & 1.4533 & 1.4251
& 0.2128 & 0.1952 & 0.1440
& 0.3599 & 0.4067 & 0.3904 \\

CAES~\citep{vader}
& 1.0840 & 0.7322 & 0.7589
& 0.5265 & 1.4070 & 1.4223
& 0.2074 & 0.1997 & 0.1506
& 0.3533 & 0.4023 & 0.3947 \\

Temporal Q-Former (Generic Prompt)
& 1.1702 & 0.9363 & 0.7610
& 0.5266 & 1.5137 & \textbf{1.5763}
& 0.2050 & 0.2086 & 0.1577
& 0.3532 & 0.4145 & \textbf{0.4050} \\

Temporal Q-Former (Object Tags)
& \textbf{1.2050} & \textbf{0.9564} & \textbf{0.7817}
& \textbf{0.5657} & \textbf{1.5745} & 1.5247
& 0.2081 & \textbf{0.2137} & \textbf{0.1600}
& 0.3596 & \textbf{0.4201} & 0.3987 \\


\bottomrule
\end{tabular}
}
\end{table*}
\begin{table*}[h]
\centering
\caption{
\textbf{C/E/V breakdown of sampler-conditioning ablations.}
This table reports clip-, event-, and video-level results for the Temporal Q-Former variants with different conditioning signals.
Text denotes using the object-tag string as sampler text context, while CLIP Embedding denotes encoding individual object tags with a CLIP text encoder, mapping them through a trainable semantic projection layer, and using the projected embeddings as the sampler conditioning sequence.
}
\label{tab:conditioning_ablation_full}
\resizebox{\textwidth}{!}{
\begin{tabular}{lcccccccccccc}
\toprule
\multirow{2}{*}{Sampler Conditioning}
& \multicolumn{3}{c}{BLEU~($\uparrow$)}
& \multicolumn{3}{c}{CIDEr~($\uparrow$)}
& \multicolumn{3}{c}{METEOR~($\uparrow$)}
& \multicolumn{3}{c}{ROUGE~($\uparrow$)} \\
\cmidrule(lr){2-4}
\cmidrule(lr){5-7}
\cmidrule(lr){8-10}
\cmidrule(lr){11-13}
& C & E & V
& C & E & V
& C & E & V
& C & E & V \\
\midrule
Generic Prompt
& 1.1702 & 0.9363 & 0.7610
& 0.5266 & 1.5137 & 1.5763
& 0.2050 & 0.2086 & 0.1577
& 0.3532 & 0.4145 & 0.4050 \\

Object Tags (Text)
& 1.1845 & 0.9417 & 0.7558
& 0.5302 & 1.5584 & \textbf{1.6293}
& 0.2044 & 0.2117 & 0.1581
& 0.3556 & \textbf{0.4204} & \textbf{0.4063} \\

Object Tags (CLIP Embedding)
& \textbf{1.2050} & \textbf{0.9564} & \textbf{0.7817}
& \textbf{0.5657} & \textbf{1.5745} & 1.5247
& \textbf{0.2081} & \textbf{0.2137} & \textbf{0.1600}
& \textbf{0.3596} & 0.4201 & 0.3987 \\


\bottomrule
\end{tabular}
}
\end{table*}

\subsection{Task-wise Sampling Analysis}
\label{app:taskwise_sampling}

To further understand this behavior, we analyze score differences by official HIVAU task labels.
HIVAU-70K provides task labels including captioning, description, analysis, and judgement across clip-, event-, and video-level samples.
For each granularity/task pair, we compute metric differences relative to ATS; positive values indicate improvement over ATS.

\begin{figure*}[h]
    \centering
    \begin{subfigure}{0.49\textwidth}
        \centering
        \includegraphics[width=\linewidth]{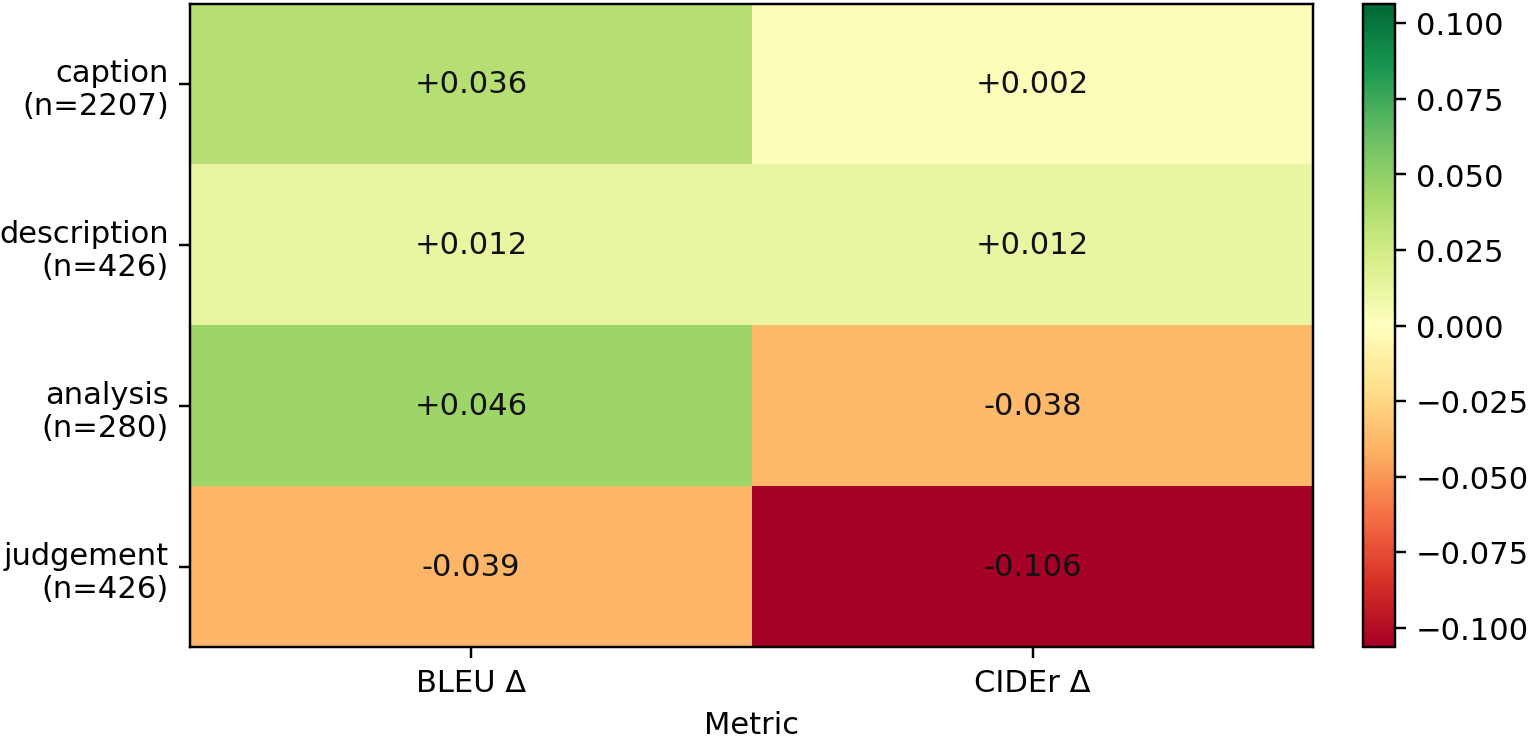}
        \caption{Uniform sampling $-$ ATS}
        \label{fig:taskwise_uniform_vs_ats}
    \end{subfigure}
    \hfill
    \begin{subfigure}{0.49\textwidth}
        \centering
        \includegraphics[width=\linewidth]{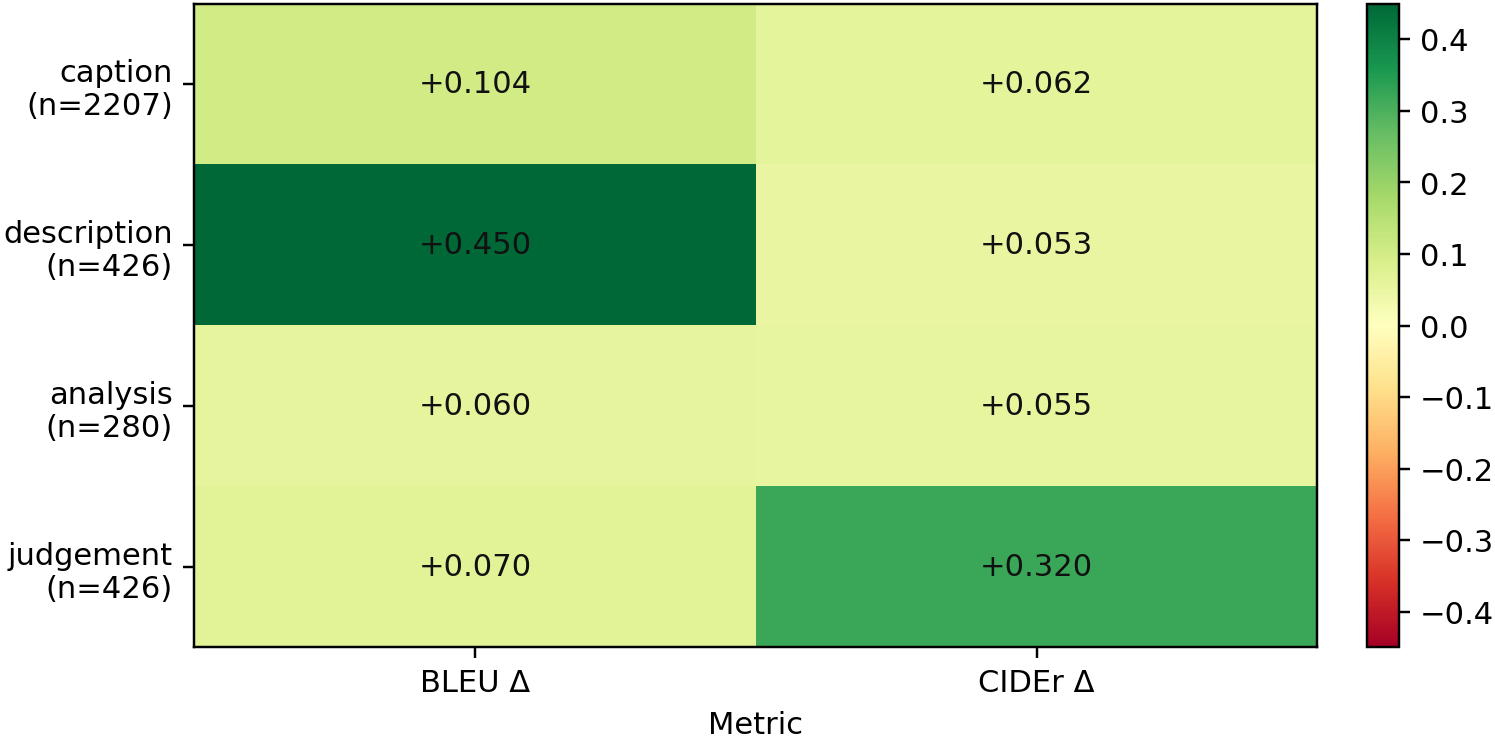}
        \caption{Temporal Q-Former w/ object tags $-$ ATS}
        \label{fig:taskwise_qformer_vs_ats}
    \end{subfigure}
    \caption{
    \textbf{Task-wise sampling analysis on HIVAU-70K.}
    Each cell reports the corpus-level metric difference over ATS for a given granularity/task pair; positive values indicate improvement over ATS.
    Uniform sampling and ATS share the same checkpoint and differ only in the inference-time sampling rule.
    The Temporal Q-Former comparison uses a separately trained end-to-end variant, so it should be interpreted as the effect of replacing the fixed sampler with a learned evidence selector.
    }
    \label{fig:taskwise_sampling}
\end{figure*}

Figure~\ref{fig:taskwise_uniform_vs_ats} compares Uniform and ATS.
Since Uniform and ATS are evaluated under the same backbone, split, and evaluation implementation, this comparison reflects the effect of fixed temporal sampling strategies.
Uniform sampling is competitive with ATS for clip-level captioning and event-level description, suggesting that broad temporal coverage can be effective when the response mainly summarizes short or descriptive content.
However, Uniform sampling performs worse on analysis and judgement tasks, especially at event/video levels, where response-critical evidence may be sparse and temporally localized.

Figure~\ref{fig:taskwise_qformer_vs_ats} compares the object-tag-conditioned Temporal Q-Former with ATS.
This comparison should be interpreted as a full-system comparison, because the Temporal Q-Former is trained end-to-end with the VAU generation objective rather than applied as a test-time-only sampling rule.
The learned sampler improves over ATS across a broader range of task types, especially for event-level description/analysis and video-level judgement.
This supports the use of learnable temporal evidence selection for heterogeneous VAU responses.

\subsection{Paired Bootstrap Confidence Intervals}
\label{app:paired_bootstrap}

We report paired bootstrap confidence intervals for the main ablation comparisons in Table~\ref{tab:paired_bootstrap}.
For each comparison, we compute
\begin{equation}
    \Delta = \mathrm{Score}(A) - \mathrm{Score}(B),
\end{equation}
where positive values indicate that method $A$ outperforms method $B$.
For each bootstrap resample, we apply the same sampled evaluation units to both methods and recompute the corpus-level metrics with the same evaluation script.
We report the 95\% confidence interval of $\Delta$.
The intervals quantify evaluation-set sampling variability and do not capture training-seed variance.

\begin{table*}[h]
\centering
\caption{
\textbf{Paired bootstrap confidence intervals for ablation comparisons.}
Each cell reports $\Delta=\mathrm{Score}(A)-\mathrm{Score}(B)$ with a 95\% confidence interval.
Positive values indicate that method $A$ outperforms method $B$.
$^\ast$ denotes that the 95\% confidence interval does not include zero.
``Prompt'' denotes Temporal Q-Former with generic-prompt conditioning, and ``Object Tags'' denotes Temporal Q-Former with CLIP-embedded object-tag conditioning.
}
\label{tab:paired_bootstrap}
\setlength{\tabcolsep}{3.5pt}
\resizebox{\textwidth}{!}{
\begin{tabular}{lcccc}
\toprule
Comparison $(A-B)$
& BLEU $\Delta$ [95\% CI]
& CIDEr $\Delta$ [95\% CI]
& METEOR $\Delta$ [95\% CI]
& ROUGE $\Delta$ [95\% CI] \\
\midrule
Object Tags $-$ Prompt
& $+0.0254~[+0.0053,+0.0449]^{\ast}$
& $+0.0320~[-0.0009,+0.0632]$
& $+0.0033~[+0.0012,+0.0053]^{\ast}$
& $+0.0047~[+0.0009,+0.0083]^{\ast}$ \\

Object Tags $-$ CAES
& $+0.1630~[+0.1130,+0.2135]^{\ast}$
& $+0.0753~[+0.0287,+0.1255]^{\ast}$
& $+0.0039~[+0.0014,+0.0068]^{\ast}$
& $+0.0085~[+0.0037,+0.0131]^{\ast}$ \\

Object Tags $-$ ATS
& $+0.2337~[+0.1755,+0.2863]^{\ast}$
& $+0.0797~[+0.0356,+0.1255]^{\ast}$
& $+0.0020~[-0.0009,+0.0049]$
& $+0.0038~[-0.0003,+0.0082]$ \\

Object Tags $-$ Uniform
& $+0.2025~[+0.1395,+0.2549]^{\ast}$
& $+0.0938~[+0.0501,+0.1395]^{\ast}$
& $+0.0001~[-0.0029,+0.0028]$
& $+0.0032~[-0.0009,+0.0075]$ \\

\bottomrule
\end{tabular}
}
\end{table*}

The paired analysis shows that the object-tag-conditioned Temporal Q-Former yields stable BLEU improvements over fixed sampling strategies.
The object-tag-conditioned variant also improves over ATS and CAES on CIDEr with confidence intervals that exclude zero.
Compared with generic-prompt conditioning, object-tag conditioning gives a stable improvement on BLEU, METEOR, and ROUGE, while the CIDEr interval lies close to the zero boundary.
METEOR and ROUGE differences are generally smaller than BLEU and CIDEr; however, the Object Tags $-$ CAES comparison remains positive across all four metrics.

\subsection{Additional Qualitative Visualization}
\label{app:additional_qualitative}
\begin{figure*}[t]
    \centering
    \includegraphics[width=\textwidth]{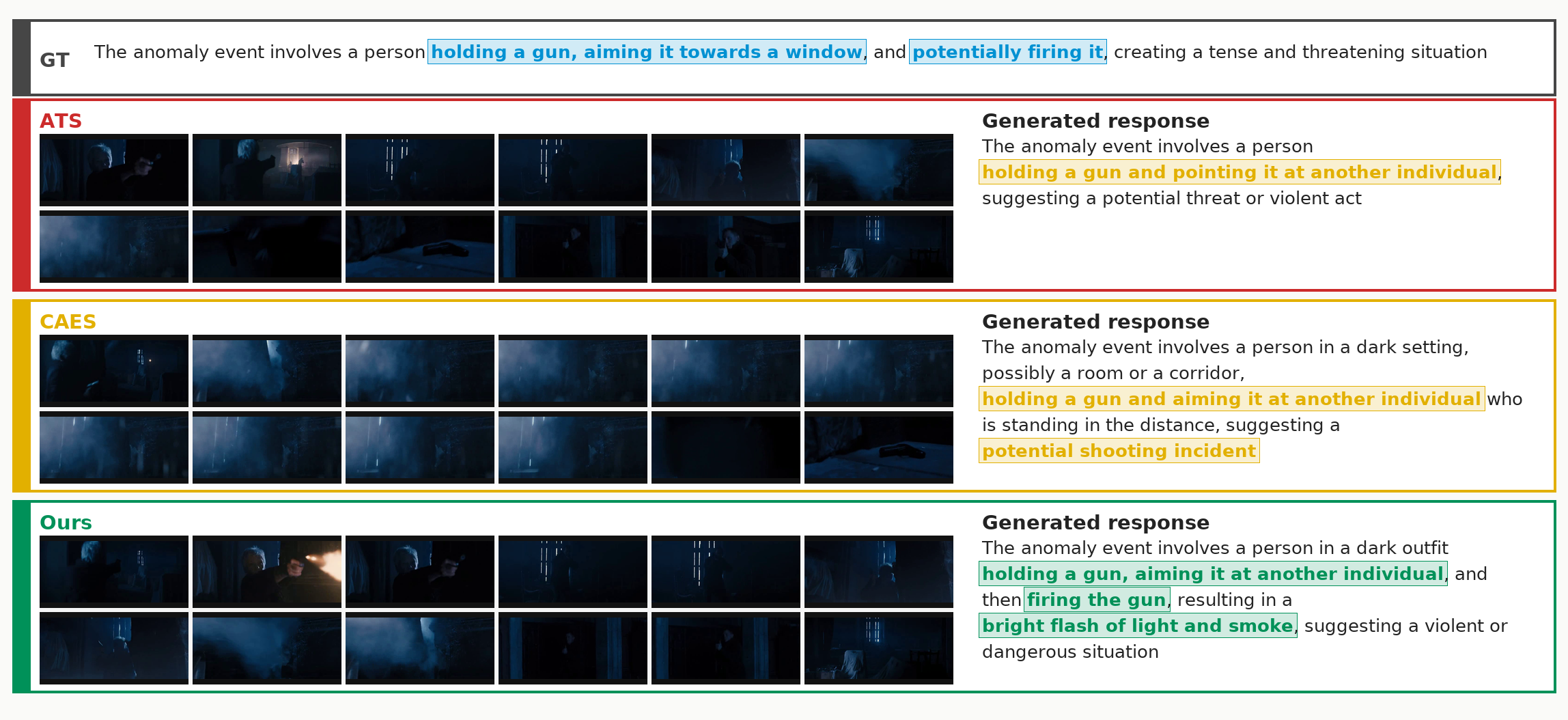}
    \includegraphics[width=\textwidth]{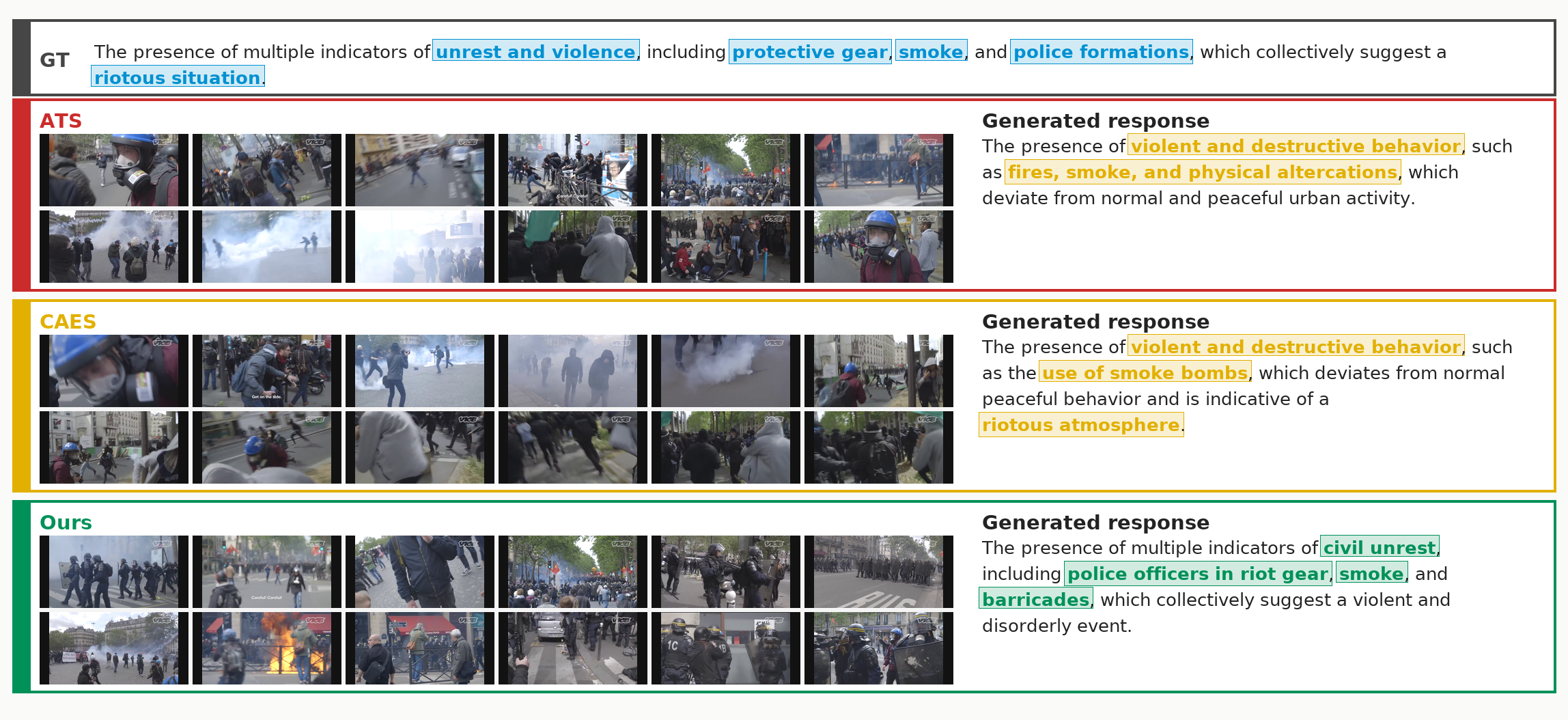}
    \includegraphics[width=\textwidth]{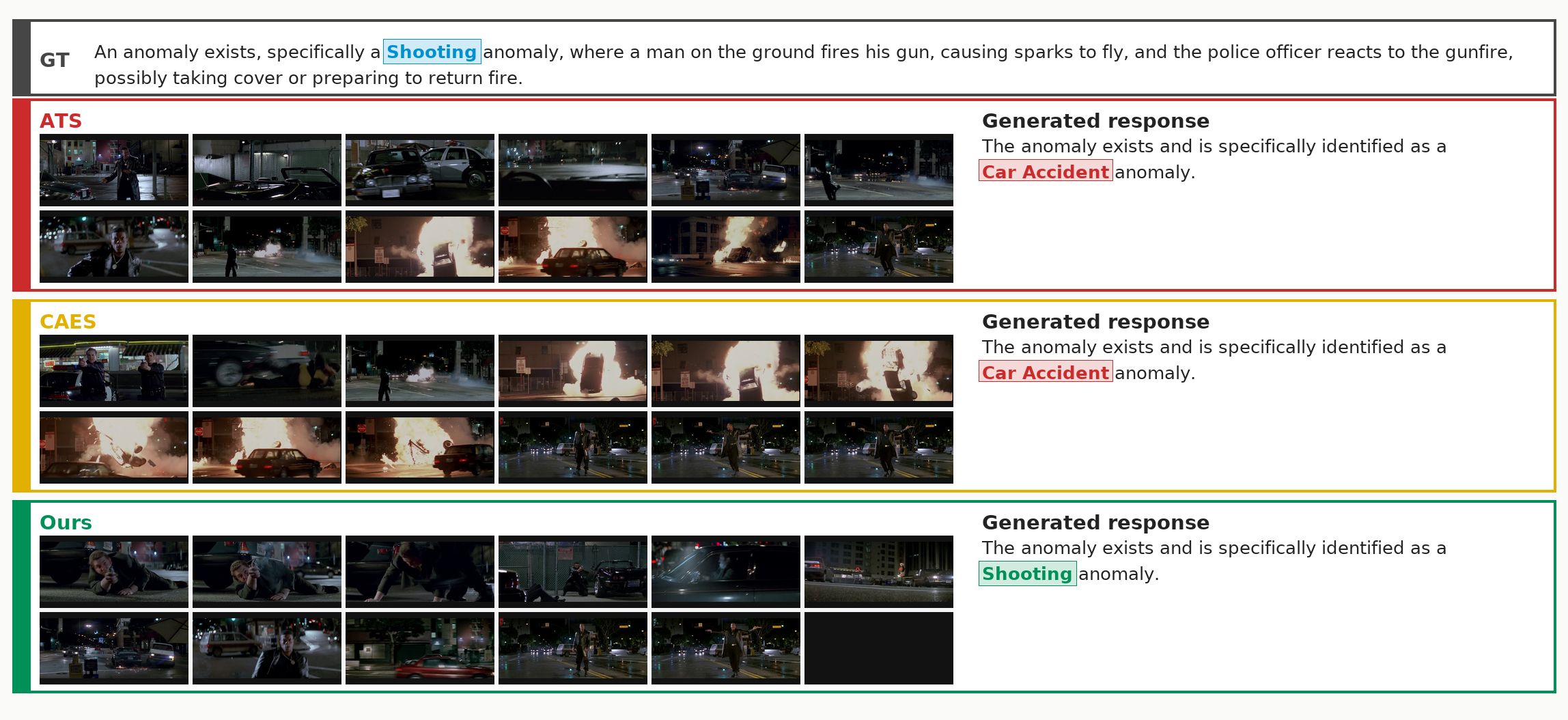}
\vspace{-5mm}
\caption{
Full sampled-frame sequences visualization for Fig.~\ref{fig:qualitative_selection}.
}
\vspace{-5mm}
\label{fig:appendix_full_sampling}
\end{figure*}

Figures~\ref{fig:appendix_full_sampling} present the full sampled-frame sequences corresponding to Fig.~\ref{fig:qualitative_selection}(a), Fig.~\ref{fig:qualitative_selection}(b), and Fig.~\ref{fig:qualitative_selection}(c), respectively. While the main paper visualizes only four representative frames per method for readability, these appendix figures show all twelve frames selected by each method. These visualizations provide a complete view of the temporal evidence used to generate the responses.

\section{Task-aware Conditioning Extension}
\label{app:task_aware_extension}

We additionally evaluate a task-aware extension of object-tag conditioning.
The motivation is that video-level object tags are invariant across prompts for the same video, whereas VAU responses may require different response intents such as description, analysis, and judgement.
The extension prepends a prompt-type tag embedding to the object-tag embedding sequence before conditioning the temporal Q-Former. Implementation details for task-aware extension are provided in Appendix~\ref{app:semantic_hint_details}.

\begin{table*}[h]
\centering
\caption{
\textbf{Comparison between Ours and Task-aware conditioning extension.}
This table reports clip-, event-, video- and full evaluation results for Ours and the task-aware conditioning extension.
}
\label{tab:task-aware_conditioning_extension}
\resizebox{\textwidth}{!}{
\begin{tabular}{lcccccccccccccccc}
\toprule
\multirow{2}{*}{Sampler Conditioning}
& \multicolumn{4}{c}{BLEU~($\uparrow$)}
& \multicolumn{4}{c}{CIDEr~($\uparrow$)}
& \multicolumn{4}{c}{METEOR~($\uparrow$)}
& \multicolumn{4}{c}{ROUGE~($\uparrow$)} \\
\cmidrule(lr){2-5}
\cmidrule(lr){6-9}
\cmidrule(lr){10-13}
\cmidrule(lr){14-17}
& C & E & V & Overall
& C & E & V & Overall
& C & E & V & Overall
& C & E & V & Overall\\
\midrule
Object Tags (CLIP Embedding)
& 1.2050 & 0.9564 & 0.7817 & \textbf{1.1482}
& 0.5657 & 1.5745 & 1.5247 & 0.9334
& 0.2081 & 0.2137 & 0.1600 & \textbf{0.2028}
& 0.3596 & 0.4201 & 0.3987 & 0.3776 \\

Task+Object Tags (CLIP Embedding)
& 1.1977 & 0.9816 & 0.7990 & 1.1413
& 0.5657 & 1.6147 & 1.5797 & \textbf{0.9488}
& 0.2064 & 0.2139 & 0.1635 & 0.2020
& 0.3596 & 0.4246 & 0.4090  & \textbf{0.3798} \\

\bottomrule
\end{tabular}
}
\end{table*}

This extension is diagnostic and is not used as our final model.
On the full evaluation set, Task+Object Tags does not provide a consistent improvement over Object Tags (CLIP Embedding), mainly because clip-level samples in HIVAU-70K are caption-only and therefore provide little task variation.
However, the C/E/V breakdown in Table~\ref{tab:task-aware_conditioning_extension} shows that Task+Object Tags improves several event- and video-level metrics, where different response intents are more meaningful.
These results suggest that prompt-type conditioning is promising for multi-intent VAU settings, but a simple global task tag is insufficient to uniformly improve full-set performance.

\section{Implementation Details}
\label{app:implementation_details}

\subsection{Training and Computation Details}
\label{app:training_compute}

All ablations use the official HIVAU-70K training split.
We train our model with \texttt{adamw} optimizer. 
The learning rate is set to $4\times10^{-5}$ with weight decay 0.01, a cosine learning-rate scheduler, and a warmup ratio of 0.03. 
We use a per-device batch size of 2 on three NVIDIA RTX 4090 24GB GPUs with gradient accumulation of 64, resulting in an effective global batch size of 384. 
Training uses BF16 mixed precision.

For parameter-efficient adaptation, we apply LoRA only to the LLM, not to the vision tower. 
The LoRA rank is 64, the LoRA scaling factor is 128, and the LoRA dropout is 0.05. 
The trainable components are the temporal Q-Former, the semantic projection layer, and the LLM LoRA adapters. 
The visual encoder, the visual-token projector, RAM tagger, CLIP text encoder, and base LLM parameters are frozen. 
Training our final model takes approximately 10 hours, on three RTX 4090 GPUs.

\subsection{Code, Data, and Existing Assets}
\label{app:code_data_assets}

We use the public HIVAU-70K benchmark with its official split. 
Our implementation relies on publicly available pretrained assets, including InternVL2-2B, RAM, and CLIP. 
The licenses of the assets used in this work are as follows: HIVAU-70K is released under the MIT license, InternVL2-2B is available on Hugging Face under the MIT license, RAM is released under the Apache License 2.0, and CLIP is released under the MIT license. 
We follow the terms of the corresponding resources and cite the original works in the main paper.

\subsection{Semantic Hint Extraction}
\label{app:semantic_hint_details}

We construct semantic hints from object-centric tags.
Object-centric tags are extracted from sparsely sampled frames using RAM~\citep{ram}.
Frames are sampled at 2 fps, with a maximum of 48 frames per video; if more than 48 frames are obtained, we uniformly resample 48 frames over the full video duration.
We use the class-wise thresholds from the official RAM implementation and aggregate frame-level tags into a video-level tag set by removing duplicates and sorting tags by occurrence frequency.

For the CLIP embedding variant used in our final model, each object tag is independently encoded using the \texttt{openai/clip-vit-large-patch14} text encoder.
Since the CLIP embedding dimension differs from the temporal Q-Former hidden dimension, each object-tag embedding is mapped by a trainable semantic projection layer.
The projected object-tag embeddings are concatenated and used as the conditioning sequence for the temporal Q-Former.

For the task-aware extension in Appendix~\ref{app:task_aware_extension}, we additionally use a task-type tag associated with the input prompt.
In HIVAU-70K, this prompt type is available from the official task metadata rather than parsed from raw prompt text.
The dataset contains four task labels: caption, description, analysis, and judgement.
Because caption samples appear only at the clip level, we map caption to description and use three prompt-type tags: description, analysis, and judgement.
The task tag is encoded with the same frozen CLIP text encoder and mapped through the same trainable semantic projection layer as the object tags.
The projected task tag embedding is prepended to the projected object-tag embeddings.


\newpage
\section*{NeurIPS Paper Checklist}

\begin{enumerate}

\item {\bf Claims}
    \item[] Question: Do the main claims made in the abstract and introduction accurately reflect the paper's contributions and scope?
    \item[] Answer: \answerYes{}.
    \item[] Justification: The abstract and introduction state the main scope of the paper: learnable temporal evidence selection for VAU and semantic conditioning via object-centric cues. The claims are supported by ablations in Section~\ref{sec:experiments}, while comparisons are explicitly separated from architectural claims.

\item {\bf Limitations}
    \item[] Question: Does the paper discuss the limitations of the work performed by the authors?
    \item[] Answer: \answerYes{}.
    \item[] Justification: The limitations and future work are discussed in Appendix~\ref{app:limitations_impacts}, including evaluation scope, dependence on off-the-shelf tag/CLIP encoders, robustness to noisy tags, and computational overhead.

\item {\bf Theory assumptions and proofs}
    \item[] Question: For each theoretical result, does the paper provide the full set of assumptions and a complete (and correct) proof?
    \item[] Answer: \answerNA{}.
    \item[] Justification: The paper does not present theoretical results, theorems, or proofs. The contribution is methodological and empirical.

\item {\bf Experimental result reproducibility}
    \item[] Question: Does the paper fully disclose all the information needed to reproduce the main experimental results of the paper to the extent that it affects the main claims and/or conclusions of the paper (regardless of whether the code and data are provided or not)?
    \item[] Answer: \answerYes{}.
    \item[] Justification: The paper describes the architecture, training objective, trainable and frozen modules, temporal sampling configuration, tag extraction procedure, evaluation metrics, and ablation protocol in Sections~\ref{sec:method}--\ref{sec:experiments} and Appendix~\ref{app:training_compute}.

\item {\bf Open access to data and code}
    \item[] Question: Does the paper provide open access to the data and code, with sufficient instructions to faithfully reproduce the main experimental results, as described in supplemental material?
    \item[] Answer: \answerNo{}.
    \item[] Justification: The current anonymous submission does not include a public code release because the implementation requires additional cleanup and documentation. The paper uses public benchmark data and public pretrained assets, and we plan to release cleaned code after acceptance.

\item {\bf Experimental setting/details}
    \item[] Question: Does the paper specify all the training and test details (e.g., data splits, hyperparameters, how they were chosen, type of optimizer) necessary to understand the results?
    \item[] Answer: \answerYes{}.
    \item[] Justification: The paper specifies the official split, evaluation metrics, candidate frame budget, number of temporal queries, Gumbel-Softmax temperature, trainable and frozen modules, optimizer, learning rate, weight decay, scheduler, warmup ratio, batch size, LoRA configuration, precision, and training steps in Section~\ref{sec:experiments} and Appendix~\ref{app:training_compute}.

\item {\bf Experiment statistical significance}
    \item[] Question: Does the paper report error bars suitably and correctly defined or other appropriate information about the statistical significance of the experiments?
    \item[] Answer: \answerYes{}.
    \item[] Justification: We report paired bootstrap confidence intervals for the main ablation comparisons in Appendix~\ref{app:paired_bootstrap}. The intervals are computed by resampling evaluation units and recomputing corpus-level metrics with the same evaluation script. They quantify evaluation-set sampling variability under the reported evaluation runs and do not capture training-seed variance.

\item {\bf Experiments compute resources}
    \item[] Question: For each experiment, does the paper provide sufficient information on the computer resources (type of compute workers, memory, time of execution) needed to reproduce the experiments?
    \item[] Answer: \answerYes{}.
    \item[] Justification: Appendix~\ref{app:training_compute} reports the GPU type, number of GPUs, memory, effective batch size, optimizer steps, precision, and wall-clock training time for the final model.

\item {\bf Code of ethics}
    \item[] Question: Does the research conducted in the paper conform, in every respect, with the NeurIPS Code of Ethics \url{https://neurips.cc/public/EthicsGuidelines}?
    \item[] Answer: \answerYes{}.
    \item[] Justification: The work uses existing benchmark data and public pretrained models for research on video anomaly understanding. The paper does not introduce human-subject experiments or collect new private data, and the potential dual-use implications of anomaly understanding are discussed.

\item {\bf Broader impacts}
    \item[] Question: Does the paper discuss both potential positive societal impacts and negative societal impacts of the work performed?
    \item[] Answer: \answerYes{}.
    \item[] Justification: Broader impacts are discussed in Appendix~\ref{app:limitations_impacts}, including potential benefits for interpretable video analysis and risks in surveillance or monitoring contexts.

\item {\bf Safeguards}
    \item[] Question: Does the paper describe safeguards that have been put in place for responsible release of data or models that have a high risk for misuse (e.g., pre-trained language models, image generators, or scraped datasets)?
    \item[] Answer: \answerNA{}.
    \item[] Justification: The paper does not release a new dataset, pretrained foundation model, or deployed system with high-risk misuse potential. If code or checkpoints are released after acceptance, they will be accompanied by appropriate usage notes and licensing information.

\item {\bf Licenses for existing assets}
    \item[] Question: Are the creators or original owners of assets (e.g., code, data, models), used in the paper, properly credited and are the license and terms of use explicitly mentioned and properly respected?
    \item[] Answer: \answerYes{}.
    \item[] Justification: The paper cites the existing benchmark, pretrained models, and metrics used in the experiments. Appendix~\ref{app:code_data_assets} lists the licenses for HIVAU-70K, InternVL2-2B, RAM, and CLIP, and states that the work follows the corresponding access terms.

\item {\bf New assets}
    \item[] Question: Are new assets introduced in the paper well documented and is the documentation provided alongside the assets?
    \item[] Answer: \answerNA{}.
    \item[] Justification: The paper does not introduce a new dataset or benchmark asset. The proposed model components are described in the method section; code release is planned after acceptance following cleanup and documentation.

\item {\bf Crowdsourcing and research with human subjects}
    \item[] Question: For crowdsourcing experiments and research with human subjects, does the paper include the full text of instructions given to participants and screenshots, if applicable, as well as details about compensation (if any)?
    \item[] Answer: \answerNA{}.
    \item[] Justification: The paper does not involve crowdsourcing experiments or new research with human subjects.

\item {\bf Institutional review board (IRB) approvals or equivalent for research with human subjects}
    \item[] Question: Does the paper describe potential risks incurred by study participants, whether such risks were disclosed to the subjects, and whether Institutional Review Board (IRB) approvals (or an equivalent approval/review based on the requirements of your country or institution) were obtained?
    \item[] Answer: \answerNA{}.
    \item[] Justification: The paper does not involve new human-subject data collection, crowdsourcing, or user studies, so IRB approval is not applicable.

\item {\bf Declaration of LLM usage}
    \item[] Question: Does the paper describe the usage of LLMs if it is an important, original, or non-standard component of the core methods in this research? Note that if the LLM is used only for writing, editing, or formatting purposes and does \emph{not} impact the core methodology, scientific rigor, or originality of the research, declaration is not required.
    \item[] Answer: \answerYes{}.
    \item[] Justification: The paper uses an LLM/MLLM as a core component for VAU response generation; the architecture, frozen/trainable modules, and LoRA-based training setup are described in Section~\ref{sec:method}. Separately, AI assistance was used for manuscript preparation, including language editing and the creation/refinement of the conceptual motivation illustration in Figure~\ref{fig:intro_motivation}. This illustration is not used as experimental evidence, training data, evaluation data, or part of the proposed method.
\end{enumerate}

\end{document}